%% file: saber.tex
\documentclass[letterpaper]{article} 
\usepackage[draft]{aaai2026}  
\usepackage{graphicx}
\usepackage{subcaption}
\usepackage{times}  
\usepackage{helvet}  
\usepackage{courier}  
\usepackage[hyphens]{url}  
\usepackage{graphicx} 
\usepackage{natbib}  
\usepackage{caption} 
\usepackage{booktabs}
\usepackage{multirow}
\usepackage{color}
\usepackage{algorithm}
\usepackage{algorithmic}
\usepackage{amsmath}
\usepackage{amsfonts}

\usepackage{newfloat}
\usepackage{listings}
\DeclareCaptionStyle{ruled}{labelfont=normalfont,labelsep=colon,strut=off} 
\floatstyle{ruled}
\newfloat{listing}{tb}{lst}{}
\floatname{listing}{Listing}
\title{SABER: Switchable and Balanced Training for Efficient LLM Reasoning}

\author{
    Kai Zhao\textsuperscript{\rm 1}\equalcontrib,
    Yanjun Zhao\textsuperscript{\rm 1}\equalcontrib,
    Jiaming Song\textsuperscript{\rm 1},
    Shien He\textsuperscript{\rm 1},
    Lusheng Zhang\textsuperscript{\rm 1},\\
    Qiang Zhang\textsuperscript{\rm 1},
    Tianjiao Li\textsuperscript{\rm 1}\thanks{Corresponding author.}
}
\affiliations{
    \textsuperscript{\rm 1}Bilibili Inc.

    \{zhaokai02, zhaoyanjun01, songjiaming, heshien, zhanglusheng01, zhangqiang, litianjiao01\}@bilibili.com
}

\usepackage{bibentry}

\begin{document}

\maketitle

\begin{abstract}

Large language models (LLMs) empowered by chain-of-thought reasoning have achieved impressive accuracy on complex tasks but suffer from excessive inference costs and latency when applied uniformly to all problems. We propose SABER (Switchable and Balanced Training for Efficient LLM Reasoning), a reinforcement learning framework that endows LLMs with user‑controllable, token‑budgeted reasoning. SABER first profiles each training example’s base‑model thinking token usage and assigns it to one of the predefined budget tiers. During fine‑tuning, the model is guided by system prompts and length‑aware rewards to respect its assigned budget. In parallel, we incorporate no‑think examples to ensure the model remains reliable even when explicit reasoning is turned off. SABER further supports four discrete inference modes—NoThink, FastThink, CoreThink, and DeepThink, enabling flexible trade‑offs between latency and reasoning depth. Extensive evaluations on math reasoning (MATH, GSM8K), code generation (MBPP), and logical reasoning (LiveBench-Reasoning) demonstrate that SABER achieves high accuracy under tight budgets, graceful degradation, and effective cross-scale and cross‑domain generalization. In particular, SABER‑FastThink cuts reasoning length by \textbf{65.4\%} and yields a \textbf{3.6\%} accuracy gain compared with the base model on the MATH benchmark.

\end{abstract}








\input{Sections/Introduction}

\input{Sections/Related_work}
\input{Sections/Method}

\input{Sections/Experiment}
\input{Sections/Conclusion}





\bibliography{aaai2026}




\end{document}

%% file: Sections/Introduction.tex
\section{Introduction}
\label{introduction}

Recent advances in large language models (LLMs)~\cite{achiam2023gpt, bai2023qwen} have significantly improved their ability to handle complex reasoning tasks through explicit step-by-step thinking. Approaches such as Chain-of-Thought prompting~\cite{wei2023chainofthoughtpromptingelicitsreasoning} and test-time reasoning expansion~\cite{test-time-compute} allow models to break down problems into intermediate steps before producing final answers. These strategies have proven effective across domains. However, they also introduce new challenges: reasoning traces tend to become excessively long, leading to inflated inference costs and latency. More critically, such reasoning behaviors are often applied uniformly across all problems, regardless of task complexity or user preferences.

This mismatch between reasoning depth and task requirement has led to what is increasingly recognized as the \emph{overthinking problem}, where LLMs generate unnecessarily elaborate reasoning even for trivial inputs~\cite{aggarwal2025l1controllinglongreasoning, han2025tokenbudgetawarellmreasoning}. For instance, instead of directly answering “What is 1 + 1?”, some models may explore multiple addition strategies, include irrelevant justifications, and consume tens of times more tokens than needed. This not only slows down response but also increases serving costs, limiting deployment efficiency. Although previous works~\cite{aggarwal2025l1controllinglongreasoning,li2025selfbudgeteradaptivetokenallocation,kimiteam2025kimik15scalingreinforcement} have attempted to shorten outputs using instruction tuning, response length control, or reward shaping, these methods typically enforce rigid constraints or rely on task-agnostic heuristics. They lack the ability to dynamically adjust reasoning length based on problem difficulty, or to give users explicit control over the reasoning process.

\begin{figure*}[t]
\centering
  \includegraphics[width=6.8in]{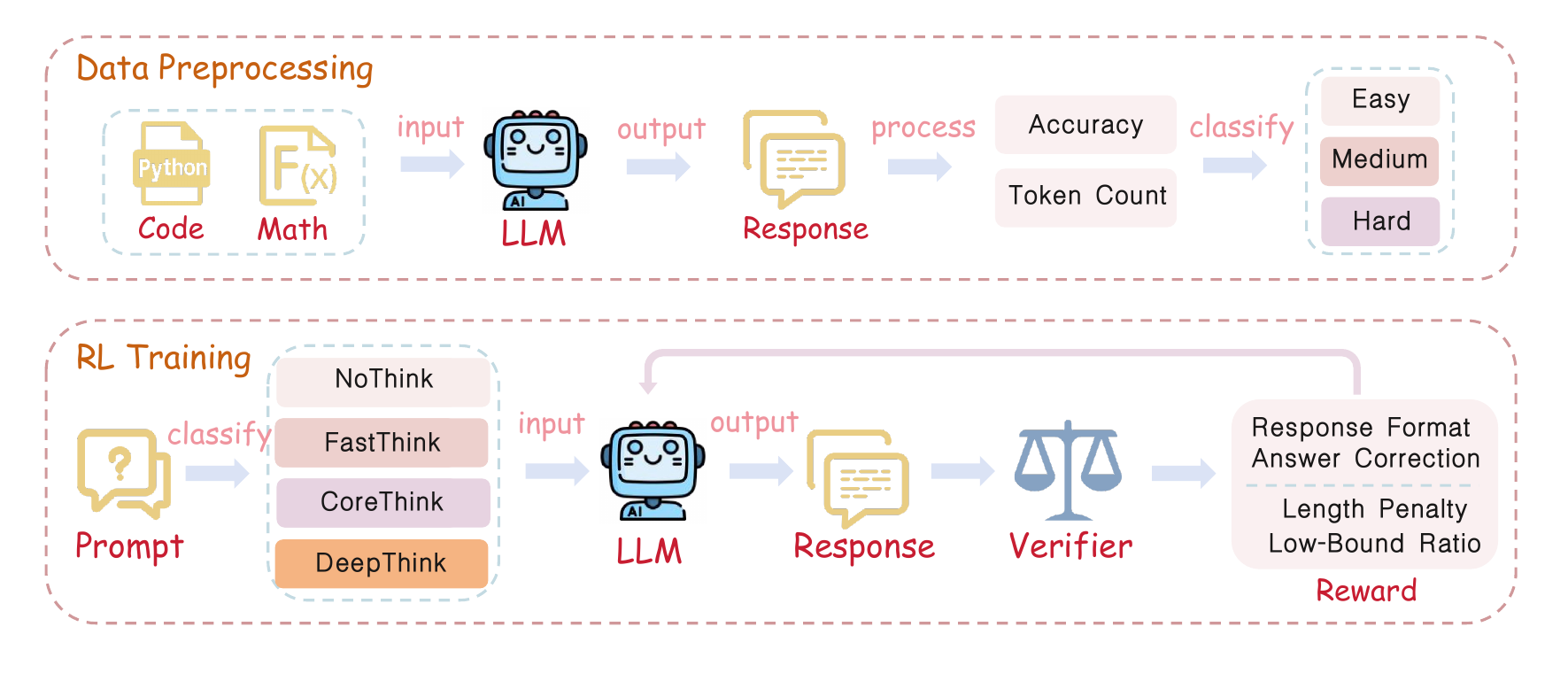}
  \vspace{-6mm}
  \caption{Overview of the SABER framework. The upper part illustrates the data preprocessing pipeline, where reasoning budget is estimated via base model inference and used to categorize training samples by difficulty (Easy / Medium / Hard). The lower part shows the RL training process, where mode-specific prompts guide the model to produce responses, which are then evaluated with multi-part rewards based on format correctness, answer accuracy, and length alignment.}
 \label{fig_structure}
 \vspace{-5mm}
\end{figure*}

To tackle these limitations, we present \textbf{SABER} — \textbf{S}witchable \textbf{A}nd \textbf{B}alanced Training for \textbf{E}fficient LLM \textbf{R}easoning, a reinforcement learning framework that enables language models to reason under explicitly specified modes. Instead of applying a uniform constraint, SABER first analyzes the base model’s output to estimate the reasoning effort required for each sample, then assigns a tiered token budget that reflects task difficulty. During RL training, the model is guided to respect its target budget using system prompts and length-aware reward shaping. This formulation encourages efficient reasoning on simple inputs while preserving long-form reasoning capability for harder examples.

In particular, SABER defines four discrete reasoning modes—\textit{NoThink}, \textit{FastThink}, \textit{CoreThink}, and \textit{DeepThink}, which allow explicit control over reasoning granularity at inference time. One key feature of SABER is its unified treatment of both thinking-enabled and thinking-disabled modes. While prior works assume reasoning is always active, real-world applications often require immediate responses without thinking process. We explicitly incorporate a curated set of thinking-disabled examples to preserve performance under this setting. As a result, SABER can gracefully support both thoughtful and direct response styles within a single model, reducing the performance gap between both thinking and no-thinking modes. Additionally, these modes are not only interpretable and user-controllable, but also generalize well across domains. Unlike previous works~\cite{li2025selfbudgeteradaptivetokenallocation, huang2025adactrladaptivecontrollablereasoning} that focus solely on the math reasoning task, SABER trains in both math reasoning and code generation tasks. We further show that SABER can scale to larger models and that the learned reasoning behaviors transfer effectively to the unseen logical reasoning task, highlighting the strong cross-scale and cross-domain generalization of SABER.

Furthermore, the training process of SABER is highly efficient and stable. While many prior RL-based efficient reasoning methods rely on supervised fine-tuning (SFT) as a necessary warm start, SABER can be directly optimized via reinforcement learning from a distilled base model. Its structure-aware design, curriculum token budget, and multipart reward formulation allow it to converge quickly without pre-training overhead. The main contributions of this work are summarized as follows:

\begin{itemize}
    \item We propose a unified framework SABER that enables efficient and stable optimization of long-thinking language models, achieving both high efficiency and training stability through budget-based data grouping, curriculum-style degradation, and reward constraints.
    
    \item SABER supports four reasoning modes — NoThink, FastThink, CoreThink, and DeepThink, allowing users to explicitly control the model's reasoning depth. This unified and switchable design accommodates diverse usage scenarios, from minimal-latency response to complex reasoning.
    
    \item Comprehensive experiments on mathematical reasoning, code generation, and logical reasoning benchmarks demonstrate that SABER maintains strong performance under tight token budgets while enabling graceful degradation and broad generalization across tasks.
\end{itemize}

%% file: Sections/Related_work.tex
\section{Related Works}
\label{Related_work}

\textbf{Overthinking in Large Reasoning Models.} 
Several studies~\cite{sui2025stopoverthinkingsurveyefficient,feng2025efficient, hou2025thinkprune,lu2025prolonged,zhuang2025accelerating,lee2025well} pursue the idea of compressing chains-of-thought without sacrificing the logical gains brought by reinforcement-learning fine-tuning. Early work~\cite{wu2025unlockingefficientlongtoshortllm} on L2S demonstrates that merging a “long-thinker” policy with a concise “answer-only” policy can yield a single model that maintains accuracy while emitting far fewer tokens. Building on this insight, mixed distillation~\cite{chenglin-etal-2024-mixed} transfers knowledge from a powerful teacher by jointly distilling both detailed and terse rationales, then applies a length-aware RL objective so the student can decide when brevity suffices. A parallel line re-tools Direct Preference Optimization (DPO)~\cite{liu2024lengthdesensitizationdirectpreference} by factoring length directly into the preference score, enabling the model to prefer short proofs when they are correct and to fall back on longer reasoning only when necessary. These ideas converge in Kimi Long2Short~\cite{kimiteam2025kimik15scalingreinforcement}, which couples mixed distillation with length-sensitive DPO and rejection sampling, achieving multi-fold speed-ups in production chat while matching its long-form teacher on mathematical benchmarks. Finally, L1~\cite{aggarwal2025l1controllinglongreasoning} generalises the paradigm by conditioning the policy on an explicit token budget during training; at inference time the very same network can “slide” along a cost-accuracy curve, adhering to a strict cap for latency-critical use cases or relaxing it when full-precision reasoning is warranted.

\textbf{Thinking Budget in Large Reasoning Models.} 
Other works embed the notion of a thinking budget directly into the optimisation objective~\cite{shen2025dast,wu2025effectively,li2025think,jiang2025think,fan2025cothink}. SelfBudgeter~\cite{li2025selfbudgeteradaptivetokenallocation} appends a special token prompting the model to predict its own budget; an RL reward then balances correctness against any budget overruns, cutting redundant reasoning by more than half on GSM8K. Instead of self-prediction, Token-Budget-Aware LLM Reasoning (TALE)~\cite{han2025tokenbudgetawarellmreasoning} trains with supervisor-provided budgets that approximate the minimum tokens needed to solve each instance, reinforcing the habit of stopping once the answer is clear. AdaCtrl~\cite{huang2025adactrladaptivecontrollablereasoning} introduces a two-stage scheme in which the model first self-assesses problem difficulty and then maps that difficulty to an adaptive budget through a learned controller, yielding substantial speed-ups without external hints. Pushing adaptivity further, Adaptive Length Penalty (ALP)~\cite{xiang2025justthinkingefficientreasoning} modulates the penalty coefficient on-line, granting more tokens only when the observed success rate justifies extra computation; this simple extension trims a third of the tokens on code-generation tasks while slightly improving pass-@-1 accuracy. 

%% file: Sections/Method.tex
\section{Method}
\label{Method}

In this section, we describe our three-stage approach for SABER framework that supports different user-controlled reasoning modes. We first present how we extract and categorize base model's thinking token statistics into discrete budget levels. Next, we detail our sample grouping and stability control mechanisms. We then introduce the injection of no-think samples to enable direct-answer mode. Finally, we formalize our reinforcement learning objective, including reward components that jointly enforce format, answer correctness, length alignment, and anti-hacking constraints. Figure~\ref{fig_structure} summarizes the framework of SABER.

\subsection{Thinking Collection and Budget Categorization}

\begin{figure}[t]
\centering
  \includegraphics[width=3.2in]{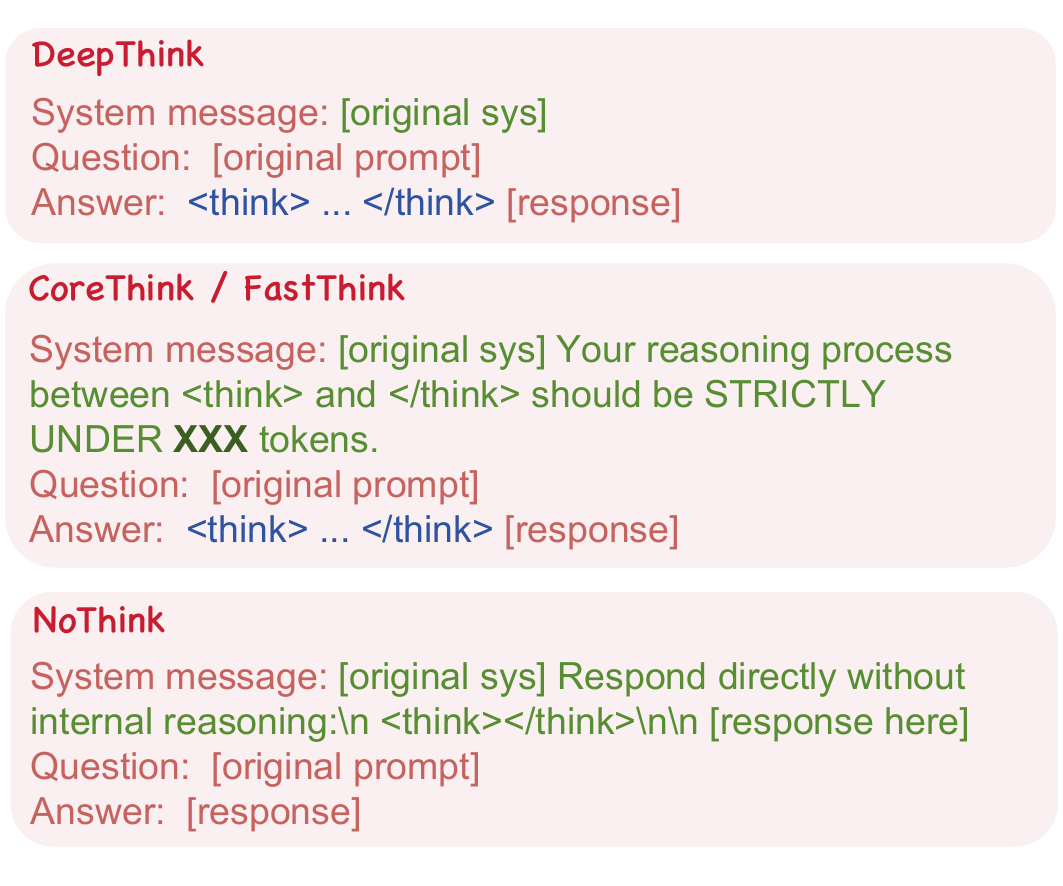}
  \vspace{-3mm}
  \caption{Templates for different modes of SABER.}
 \label{prepend_prompt}
 \vspace{-5mm}
\end{figure}

The design of thinking budget is the heart of SABER. If every example’s target budget were set to the same value, problems requiring few thinking tokens would never incur a length penalty and thus would fail to learn how to switch reasoning modes, while problems requiring many thinking tokens would be continuously penalized and quickly collapse in performance. To address this, SABER applies a per-example budget calibration. First, we run the base model over the entire training set and record the number of tokens generated between \texttt{<think>} and \texttt{</think>}. Based on the observed distribution and empirical judgments, we partition examples into three difficulty tiers—128 (easy), 4,096 (medium) and 16,384 (hard), and assign each example a corresponding target budget as follows:
\begin{itemize}
  \item If an example’s thinking token count is less than 128, we still set its target budget to 128.
  \item If it lies between 128 and 4,096, we set its target budget to 128.
  \item If it lies between 4,096 and 16,384, we set its target budget to 4,096.
  \item If it exceeds 16,384, we impose no upper bound, allowing the model full freedom of reasoning on these challenging tasks.
\end{itemize}
We then prepend each prompt with a system message, as shown in Figure~\ref{prepend_prompt}, 
where \texttt{XXX} is replaced by the example’s target budget and DeepThink mode has no target budget.

This tiered downgrade strategy offers two key advantages:
(1) it ensures that more training examples incur a length penalty from the outset, thereby efficiently teaching the model to switch between reasoning modes;
(2) it respects the inherent variation in reasoning lengths across examples, enabling smooth transitions between adjacent modes and balancing training efficiency with stability.

\subsection{Stabilizing Mode Transition with Penalties}

Applying length penalties uniformly to all training examples from the beginning of fine-tuning can introduce two key issues. First, frequent and abrupt switching between reasoning modes may destabilize training. Second, the model may become overly biased toward generating shorter reasoning traces, leading to underthinking and potential performance degradation. To mitigate these risks, we introduce two complementary strategies that enable smoother and more stable mode transitions.

\paragraph{Accuracy-Based Partitioning of Training Data.}  
We evaluate the base model's ability to answer each training sample correctly. Among the examples that base model fails to solve (approximately 40\% of the corpus), we apply two treatments:
\begin{itemize}
  \item Half are retained at their original budget level, thereby reducing their exposure to length penalties.
  \item The other half are assigned no target budget at all, allowing unrestricted reasoning during training.
\end{itemize}
Only the remaining 60\% examples the base model can answer correctly are subject to the downgrade procedure. This partitioning scheme ensures that length constraints are applied more conservatively at the early stage of training, promoting stable learning of reasoning-mode switching.

\paragraph{Lower-Bound Ratio Constraint to Prevent Reward Hacking.}
To avoid degenerate behavior where the model aggressively shortens its reasoning just to minimize penalties, we introduce an additional lower-bound constraint on the generated reasoning length. Specifically, we require that the generated think-token counts $t_{\text{gen}}$ remains within a certain proportion of the base model's counts $t_{\text{base}}$. Formally, we enforce the following constraint:
\[
0.2 \cdot t_{\text{base}} \le t_{\text{gen}} \le 1.2 \cdot t_{\text{base}}.
\]
This range ensures that the model maintains sufficient reasoning content while respecting the target budget, effectively discouraging reward hacking through excessive brevity.

\subsection{User-Controlled No-Think Mode}

In real-world applications, users may sometimes prefer to receive direct answers without any intermediate reasoning~\cite{xiang2025justthinkingefficientreasoning, chen2024think23overthinkingo1like}. However, directly disabling the reasoning component in a long-thought model without any targeted adaptation often leads to a severe drop in performance. This highlights the necessity of explicitly training the model to handle such no-think scenarios.

To this end, we augment the training corpus with a subset of specially constructed no-think examples. We find that even a modest proportion of such data substantially improves the model's compatibility with no-think mode, mitigating performance degradation when reasoning is intentionally skipped. Each no-think example is constructed by manually appending a minimal reasoning block to the input, as shown in Figure~\ref{prepend_prompt}. This format explicitly instructs the model to bypass the \texttt{<think>} span and generate a response immediately. By learning from these examples, the model acquires the ability to gracefully handle user requests for direct answers, offering greater flexibility across use cases.

\subsection{Direct RL Optimization Without SFT Warmup}

Unlike many prior approaches~\cite{huang2025blendingsupervisedreinforcementfinetuning,huang2025adactrladaptivecontrollablereasoning,li2025selfbudgeteradaptivetokenallocation,ma2025learningreinforcementlearningcant} that require supervised fine-tuning (SFT) as a warm start before reinforcement learning (RL), our SABER framework introduces operations that are naturally aligned with the model’s behavior and training dynamics. As a result, SABER does not require a separate SFT warmup stage and can be directly trained using reinforcement learning from the outset. This significantly simplifies the training pipeline and reduces computational overhead.

In this work, we adopt the widely used Group Relative Policy Optimization (GRPO) algorithm~\cite{shao2024deepseekmathpushinglimitsmathematical} to fine-tune the model with structured reward signals. The GRPO objective is defined as:

\begin{align*}
\mathcal{J}_{\mathrm{GRPO}}(\theta) =\;
& \mathbb{E}_{q \sim P(Q),\, \{o_i\}_{i=1}^G \sim \pi_{\theta_{\text{old}}}(O \mid q)} \Bigg[
\frac{1}{G} \sum_{i=1}^G \frac{1}{|o_i|} \sum_{t=1}^{|o_i|} \\
& \Bigg( \min \Bigg\{
\frac{
\pi_\theta(o_{i,t} \mid q, o_{i,<t})
}{
\pi_{\theta_{\text{old}}}(o_{i,t} \mid q, o_{i,<t})
} \hat{A}_{i,t},\; \\
& \operatorname{clip} \left(
\frac{
\pi_\theta(o_{i,t} \mid q, o_{i,<t})
}{
\pi_{\theta_{\text{old}}}(o_{i,t} \mid q, o_{i,<t})
},
1{-}\varepsilon,\; 1{+}\varepsilon
\right) \hat{A}_{i,t}
\Bigg\} \\
& \quad - \beta\, D_{\mathrm{KL}}\left( \pi_\theta \,\|\, \pi_{\text{ref}} \right) \Bigg) \Bigg]
\end{align*}

Reward design lies at the core of reinforcement learning optimization. As discussed before, the SABER framework incorporates a composite reward signal composed of four distinct components:

\[
r = r_{\mathrm{format}} + r_{\mathrm{answer}} + r_{\mathrm{length}} + r_{\mathrm{ratio}}.
\]
\begin{itemize}
  \item \textbf{Format Reward}:
  \[
    r_{\mathrm{format}} = 
    \begin{cases}
        \begin{array}{ll}
        0, & \text{if format is correct}, \\
        -1, & \text{if format is wrong}.
        \end{array}
    \end{cases}
  \]
  The format reward is designed to ensure that the model's output is syntactically parsable. Specifically, it enforces a structured format of \texttt{<think>}...\texttt{</think>}... to clearly separate the model's internal reasoning content from the final answer.
  \item \textbf{Answer Reward}:
  \[
    r_{\mathrm{answer}} = 
    \begin{cases}
        \begin{array}{ll}
        1, & \text{if answer is correct}, \\
        0, & \text{if answer is wrong}.
        \end{array}
    \end{cases}
  \]
  The answer reward evaluates the correctness of the model's final response. For mathematical problems, the predicted answer is extracted from the \texttt{\textbackslash boxed\{\}} and compared with the ground-truth label. For code generation tasks, the reward is computed by extracting the code block enclosed within triple backticks (\texttt{```}), executing it in the secure sandbox environment, and verifying it against predefined test cases.
  \item \textbf{Length Penalty}:
  \[
    r_{\mathrm{length}} = 
    \begin{cases}
        \begin{array}{ll}
        0, & \text{if } t_{\mathrm{gen}} \leq t_{\mathrm{budget}}, \\
        -0.4, & \text{otherwise}.
        \end{array}
    \end{cases}
  \]
  where $t_{\mathrm{gen}}$ is the number of thinking tokens generated by the current policy. This length penalty is set to -0.4 to distinguish it from the format reward.
  \item \textbf{Lower-Bound Ratio Penalty}:
  \[
    r_{\mathrm{ratio}} = 
    \begin{cases}
    \begin{array}{ll}
        0, & \text{if } 0.2 \cdot t_{\mathrm{base}} \leq t_{\mathrm{gen}} \leq 1.2 \cdot t_{\mathrm{base}}, \\
        -0.4, & \text{otherwise}.
        \end{array}
    \end{cases}
  \]
  where $t_{\mathrm{base}}$ denotes the number of thinking tokens generated by the base policy. This constraint prevents the model from abusing short or long generations for reward hacking.
\end{itemize}

By jointly optimizing these rewards, our method achieves precise alignment to user-selected reasoning modes, enforces smooth transitions across different modes, and maintains high answer quality in long-reasoning, short-reasoning and no-reasoning scenarios.

%% file: Sections/Experiment.tex
\section{Experiments}
\label{Experiment}

\begin{table*}[t]
\centering
\caption{Main performance comparison for 1.5B size models on math and code benchmark, showing accuracy (Acc), average response length (Len).}
\label{tab:main_results_1B}
\begin{tabular}{lccccccccccccccc} 
\toprule
\multirow{2}{*}{Model} & \multicolumn{2}{c}{MATH500} & \multicolumn{2}{c}{GSM8K} & \multicolumn{2}{c}{MATH} & \multicolumn{2}{c}{MBPP}\\
\cmidrule(lr){2-3} \cmidrule(lr){4-5} \cmidrule(lr){6-7} \cmidrule(lr){8-9}
 & Acc$\uparrow$ & Len$\downarrow$ & Acc$\uparrow$ & Len$\downarrow$ & Acc$\uparrow$ & Len$\downarrow$ & Acc$\uparrow$ & Len$\downarrow$ \\
\midrule
Deepseek-R1-Distill-Qwen-1.5B & 78.9 & 8042  & 82.1 & 3471 & 80.6 & 7866 & 38.9 & 5755\\
L1-Max (1.5B, Num Tokens=512)   & - & - & 72.9 & 569 & 71.2 & 1533 & - & -\\
L1-Max (1.5B, Num Tokens=3600)   & - & - & 74.4 & 633  &77.1 &1545 & - & -\\
SelfBudgeter (1.5B, slk, Train\_Data=90k)  & - & - & 81.5 & 662& 74.2 &919 & - & -\\

\textbf{SABER} - 1.5B - DeepThink & 83.2 & 5353 & 85.7 & 1947 &85.2 &4748 &53.7 &3010 \\
\textbf{SABER} - 1.5B - CoreThink & 82.1 & 3294 & 83.1 & 930 &84.3 &3045 &49.8 &1254 \\
\textbf{SABER} - 1.5B - FastThink & 81.4 & 2899 & 82.5 & 778 &83.5 &2719 &45.1 &942 \\

\midrule
Deepseek-R1-Distill-Qwen-1.5B - NoThink & 65.1 & 0 & 61.1 & 0 & 65.5 & 0 & 30.7 & 0\\
\textbf{SABER} - 1.5B - NoThink & 76.3 & 0 & 78.1 & 0 & 76.9 & 0 &44.7 &0 \\

\bottomrule
\end{tabular}
\end{table*}

\begin{table*}[t]
\centering
\caption{Evaluation of cross‑scale and cross‑domain generalization for SABER on math, code and logic reasoning tasks, showing accuracy (Acc), average response length (Len).}
\label{tab:main_results_7B}
\begin{tabular}{lccccccccccccccc} 
\toprule
\multirow{2}{*}{Model} & \multicolumn{2}{c}{MATH500} & \multicolumn{2}{c}{GSM8K} & \multicolumn{2}{c}{MATH} & \multicolumn{2}{c}{MBPP} & \multicolumn{2}{c}{LiveBench-R}\\
\cmidrule(lr){2-3} \cmidrule(lr){4-5} \cmidrule(lr){6-7} \cmidrule(lr){8-9} \cmidrule(lr){10-11}
 & Acc$\uparrow$ & Len$\downarrow$ & Acc$\uparrow$ & Len$\downarrow$ & Acc$\uparrow$ & Len$\downarrow$ & Acc$\uparrow$ & Len$\downarrow$ & Acc$\uparrow$ & Len$\downarrow$ \\
\midrule
Deepseek-R1-Distill-Qwen-7B & 91.5 & 8221  & 91.4 & 3402 & 92.4 & 7838 & 59.9 & 5021 & 36.4 & 5619\\
\textbf{SABER} - 7B - DeepThink & 91.9 & 6265 & 92.0 & 2351 & 92.9 & 5763 & 65.8 & 2426 & 38.3 & 4401 \\
\textbf{SABER} - 7B - CoreThink & 90.1 & 2820 & 91.7 & 1319 & 91.2 & 2557 & 63.0 & 1556 & 32.1 & 3039 \\
\textbf{SABER} - 7B - FastThink & 86.5 & 1563 & 90.3 & 495 & 87.2 & 1478 & 61.1 & 969 & 30.6 & 2166 \\

\midrule
Deepseek-R1-Distill-Qwen-7B - NoThink & 79.1 & 0  & 86.4 & 0 & 79.7 & 0 & 43.2 & 0 & 17.9 & 0\\
\textbf{SABER} - 7B - NoThink & 85.1 & 0 & 89.2 & 0 & 85.5 & 0 & 58.8 & 0 & 26.8 & 0 \\

\bottomrule
\end{tabular}
\end{table*}

In this section, we present the empirical evaluations of the SABER framework to answer the following research questions (RQs): 

\textbf{RQ1}: How does SABER compare with existing strong baselines on math reasoning and code generation tasks?

\textbf{RQ2}: Does SABER generalize to larger models and unseen reasoning domains?

\textbf{RQ3}: How important are the individual design choices in SABER?

\textbf{RQ4}: What qualitative differences emerge among the switchable reasoning modes of SABER?

To address these questions, we begin by outlining the experimental setup, including the benchmarks, baselines, and training data. We then present the main results obtained using a 1.5B model on two core tasks: math reasoning (MATH/GSM8K) and code generation (MBPP). Next, we demonstrate cross‑scale and cross‑domain generalization by applying the same training recipe to a 7B model and to a logical reasoning benchmark (LiveBench-Reasoning). In addition, we conduct ablation studies by removing each component of SABER individually. Finally, we conduct a behavioral analysis of the FastThink, CoreThink, and DeepThink modes, highlighting how each mode affects reasoning depth and answer accuracy.

\begin{table*}[t]
\centering
\caption{Ablation results of SABER's individual components on math reasoning tasks, showing accuracy (Acc), average response length (Len).}
\label{tab:ablation_result}
\begin{tabular}{lccccccccccccc} 
\toprule
\multirow{2}{*}{Training Setting} & \multirow{2}{*}{Test Setting} & \multicolumn{2}{c}{MATH500} & \multicolumn{2}{c}{GSM8K} & \multicolumn{2}{c}{MATH}\\
\cmidrule(lr){3-4} \cmidrule(lr){5-6} \cmidrule(lr){7-8}
 & & Acc$\uparrow$ & Len$\downarrow$ & Acc$\uparrow$ & Len$\downarrow$ & Acc$\uparrow$ & Len$\downarrow$ \\
\midrule
\multirow{4}{*}{All Budget Downgrade}

&DeepThink &81.7  &4902 & 84.8 & 1771 &84.2 &4579 \\
&CoreThink & 80.2 & 3663 & 82.8 & 850 &82.8 &3147 \\
&FastThink & 80.3 & 3042 & 81.0 & 674 &82.0 &2731 \\
&NoThink & 72.5 & 0 & 74.2 & 0 &75.1 &0 \\
\midrule

\multirow{4}{*}{Without Budget Downgrade}
&DeepThink & 83.3 & 5636 & 85.5 & 2182 &84.9 &5261 \\
&CoreThink & 81.7 & 5378 & 84.1 & 1763 &84.4 &4856 \\
&FastThink & 81.7 & 4896 & 83.3 & 1627 &84.4 &4583 \\
&NoThink & 75.1 & 0 & 74.3 & 0 &75.7 &0 \\
\midrule

\multirow{4}{*}{Reduce NoThink Ratio}
&DeepThink & 81.7 & 5072 & 85.2 & 1944 &84.7 &4957 \\
&CoreThink & 80.7 & 3816 & 82.7 & 977 &83.1 &3325 \\
&FastThink & 79.9 & 3424 & 81.8 & 847 &82.5 &3107 \\
&NoThink & 72.5 & 0 & 73.7 & 0 &74.7 &0 \\
\midrule

\multirow{4}{*}{Remove NoThink Data}
&DeepThink & 82.1 & 5317 & 84.9 & 1838 &84.9 &4767 \\
&CoreThink & 82.5 & 3819 & 82.7 & 1133 &83.3 &3545 \\
&FastThink & 81.0 & 3488 & 81.9 & 915 &82.6 &3154 \\
&NoThink & 61.4 & 0 & 58.2 & 0 &62.3 &0 \\
\midrule

\multirow{4}{*}{Without Accuracy Filtering}
&DeepThink & 83.7 & 5774 & 84.6 & 2210 &85.0 &5501 \\
&CoreThink & 83.0 & 4845 & 82.1 &1446 &84.2 &4361 \\
&FastThink & 81.3 & 4254 & 80.5 & 960 &83.7 &4020 \\
&NoThink & 71.4 & 0 & 70.2 & 0 &71.4 &0 \\
\midrule
\bottomrule
\end{tabular}
\end{table*}

\subsection{Experimental Setups}

\textbf{Benchmark.} We evaluate on math reasoning and code generation tasks, including the following: (1) MATH500~\cite{lightman2023lets} is a 500-problem slice of the full MATH benchmark, designed to span seven contest domains while remaining compact for quick evaluation. 
(2) GSM8K~\cite{Cobbe2021TrainingVT} consists of 8.5K crowd-sourced grade-school word problems, each solvable in a few arithmetic steps. 
(3) MATH~\cite{hendrycksmath2021} scales up to 12.5K AMC/AIME-style competition questions with full step-by-step solutions covering algebra, geometry, combinatorics, and more. 
(4) MBPP~\cite{austin2021programsynthesislargelanguage} contains 974 programming challenges designed to test LLMs’ ability to generate code that solves algorithmic problems. 
(5) LiveBench-Reasoning, a subset of LiveBench~\cite{white2024livebench}, is designed to evaluate the model’s logical reasoning ability and contains 200 complex logic puzzles.

\textbf{Baseline.} We use DeepSeek-R1-Distill-Qwen-1.5B as our base model for the main experiment. This model is specifically distilled from DeepSeek-R1 and achieves new state-of-the-art results among LLMs of similar scale. We choose the 1.5B model size for comparison, as many related works in this domain adopt the same scale, making it a standard reference point for fair evaluation. To assess the effectiveness of SABER, we also compare it against two related methods: (1) L1~\cite{aggarwal2025l1controllinglongreasoning}, which proposes length-controlled policy optimization to produce outputs that adhere to strict length constraints specified in the prompt. (2) SelfBudgeter~\cite{li2025selfbudgeteradaptivetokenallocation}, which autonomously predicts the required token budgets for reasoning and effectively adheres to self-imposed constraints. We further validate SABER on a larger 7B model. However, due to the absence of published results from the two baseline methods at this scale, we only compare against the base model, DeepSeek-R1-Distill-Qwen-7B. For the evaluation results of L1 and SelfBudgeter in our experiments, we directly use the original data reported in the paper of SelfBudgeter. For DeepSeek-R1-Distill-Qwen and SABER, we use the inference configurations recommended by DeepSeek for evaluation, including temperature, top-p, and other decoding parameters, to ensure a fair and consistent comparison across methods. Our training implementation is built on the open source verl framework~\cite{sheng2024hybridflow}, available at: \url{https://github.com/volcengine/verl}.

\textbf{Training Data.} The training of SABER uses only 2K examples: 1K math and 1K code instances. Specifically, for the math domain, we first filter samples from the OpenR1-Math-220k dataset (\url{https://huggingface.co/datasets/open-r1/OpenR1-Math-220k}) using a math verifier to retain only those verified as correct, and then randomly sample 1K instances from the verified subset as SABER’s training data. For code, we randomly sample 1K examples from the KodCode-Light-RL-10K (\url{https://huggingface.co/datasets/KodCode/KodCode-Light-RL-10K}) dataset. We train SABER for 10 epochs on this combined set of 2K examples, and the resulting model is used for all experiments presented in this work. These data are partitioned using an accuracy-based strategy to promote stable learning of reasoning mode switching. As shown in Figure~\ref{fig_structure}, each instance is first annotated with a reasoning-budget tier during data preprocessing. To incorporate the NoThink mode, we further create a duplicate of each sample annotated with a zero reasoning budget, resulting in complete overlap with the core corpus. This ensures that the model encounters both constrained and direct-answer variants within the same training schedule.

\begin{figure*}[t]
\centering
  \includegraphics[width=7in]{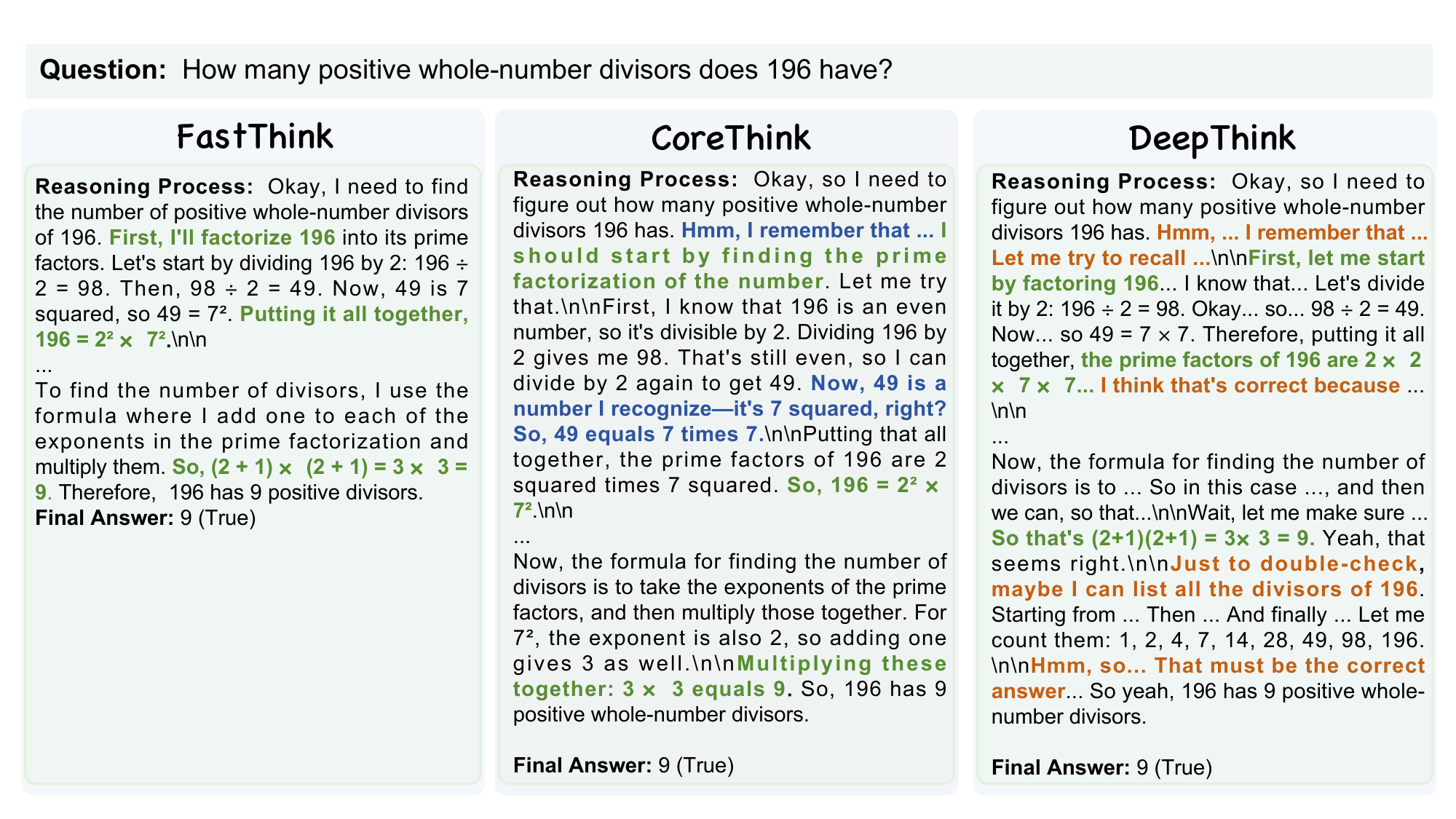}
  \caption{Given same problem, SABER supports four thinking modes. Except NoThink mode, the other three modes include the core solution steps, such as factorization and counting (shown in green). Additional reasoning and reflection unique to CoreThink are marked in blue. Further self-checking and justification steps in DeepThink are highlighted in brown.
  }
 \label{fig_case_example}
\end{figure*}

\subsection{Main Results}
To answer RQ1, Table~\ref{tab:main_results_1B} compares baselines with variants of our SABER framework, which incorporates different thinking modes. To capture both effectiveness and computational cost, we report accuracy (Acc.) and the average thinking length produced by the model (Len.). L1 effectively shortens chains of thought by rigorously capping the number of thinking tokens during training. However, this hard constraint results in a steep drop in accuracy. SelfBudgeter softens the trade-off by pre-estimating suitable budgets, recovering much of the lost performance. Nevertheless, it requires multiple training passes over 30K examples for 3 epochs, thereby incurring substantial computational cost. This inefficiency stems from the lack of task-aware budget assignment: since all samples are treated equally without token-level degradation, only a small fraction of them receive effective length penalty during training. As a result, the method requires a large number of training instances to achieve convergence, significantly increasing the overall training burden.

In contrast, SABER is trained on only 2K examples and equips the model with the ability to assess problem difficulty, demonstrating high efficiency in learning to switch reasoning modes. Compared to the base model (DeepSeek-R1-Distill-Qwen-1.5B), SABER-FastThink achieves a 72.7\% reduction in average reasoning length. Notably, this concise reasoning does not impair accuracy, on the contrary, it yields an average 3.0\% improvement across all benchmarks, demonstrating a more favorable trade-off between computational cost and performance. SABER-CoreThink further improves this balance, reducing reasoning length by 67.9\% while boosting accuracy by 4.7\%. SABER-DeepThink delivers the strongest results, cutting reasoning by 41.2\% and enhancing accuracy by 6.8\%. These gains can be attributed to two main factors. First, more concise reasoning helps avoid unnecessary repetition or distractions that may interfere with the final answer. Second, we observed that SFT-trained long-thinking models tend to exhibit repetitive generation. After reinforcement learning with format constraints, such repetition is significantly reduced, leading to improved benchmark performance.

\begin{figure*}[htbp]
\centering
  \begin{subfigure}[b]{0.45\linewidth}    
    \includegraphics[width=\linewidth]{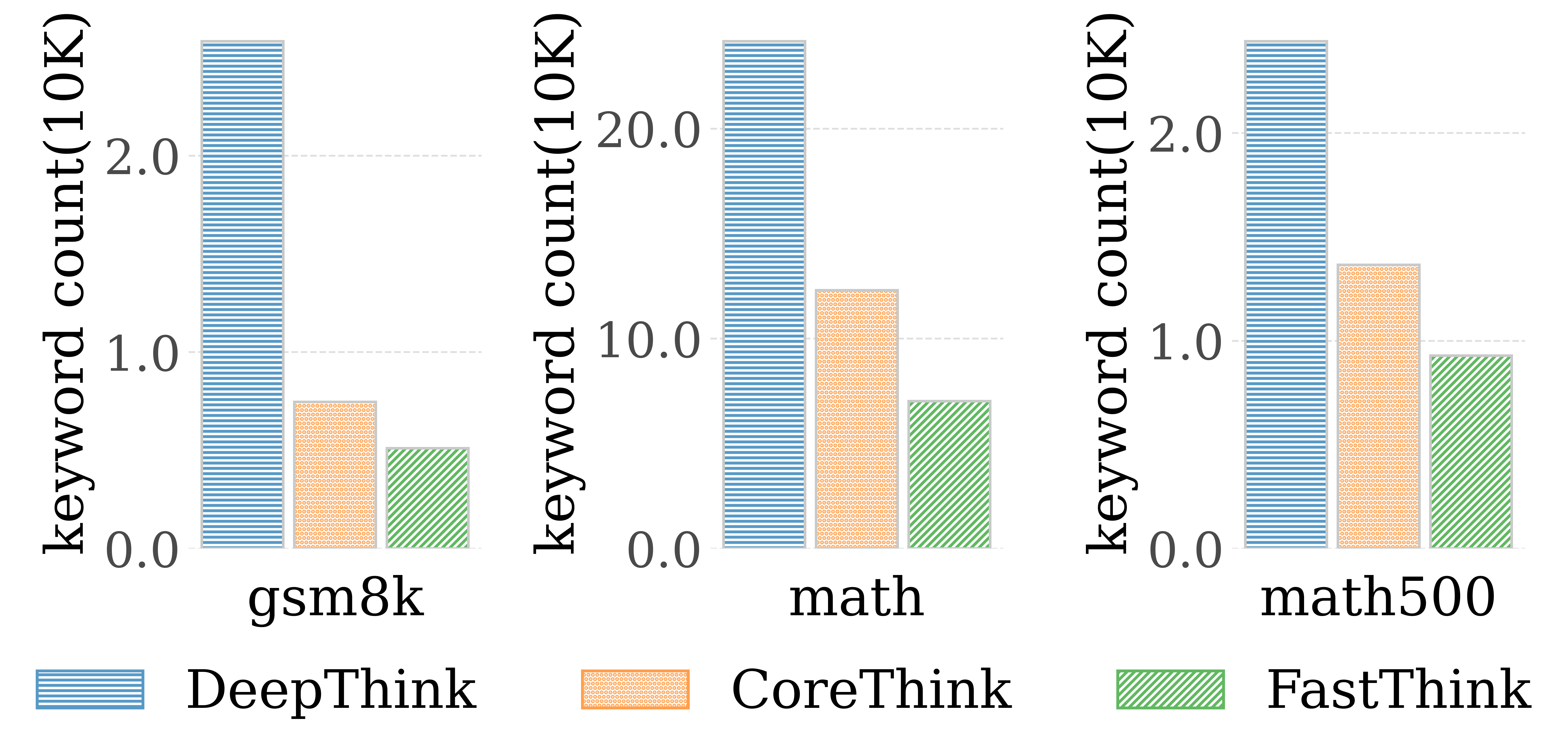}
    \caption{Frequency of reasoning-related keywords across math datasets.}
    \label{fig_freq}
  \end{subfigure}
  \hfill                       
  \begin{subfigure}[b]
  {0.45\linewidth}
    \includegraphics[width=\linewidth]{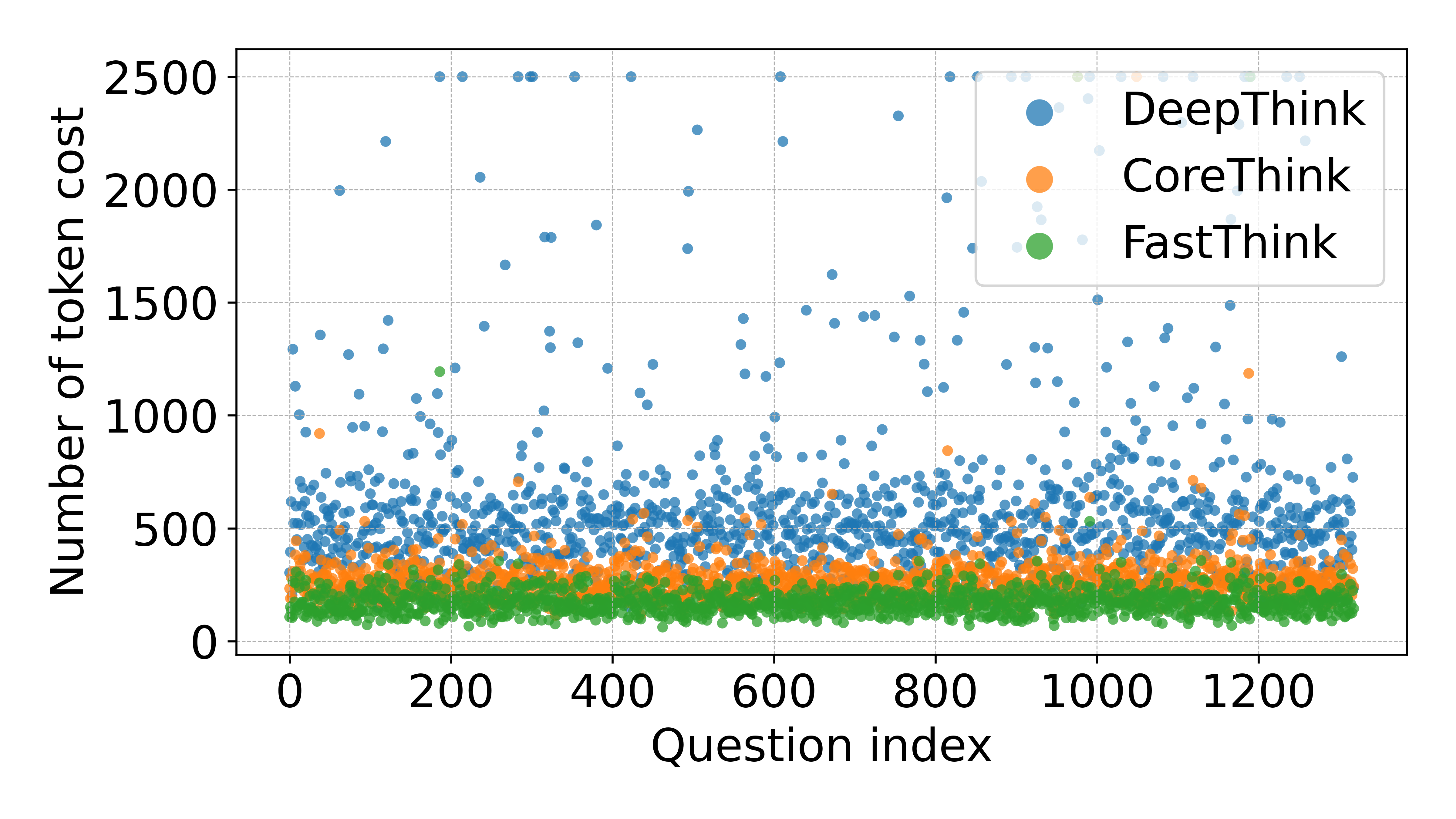}
    \caption{Distribution of thinking length under three modes on GSM8K.}
    \label{fig_cot_len}
  \end{subfigure}

   \caption{Analysis of reasoning behavior under different thinking modes. (a) Reasoning keyword frequency on GSM8K, MATH, and MATH500 shows deeper modes generate more explicit reasoning. (b) Token length distribution on GSM8K highlights clear separation in reasoning depth and inference cost among DeepThink, CoreThink, and FastThink.}
\end{figure*}

\subsection{Cross‑scale and Cross‑domain Generalization}

To answer RQ2, we further apply SABER to the larger DeepSeek-R1-Distill-Qwen-7B model to assess its cross-scale generalization. The results are shown in Table~\ref{tab:main_results_7B}. In this setting, SABER-FastThink maintains its efficiency advantage, reducing average reasoning length by a substantial 82.1\%. However, this comes with a 2.5\% decline in accuracy, reflecting a trade-off that becomes more pronounced at scale. Meanwhile, SABER-DeepThink achieves a 33.2\% reduction in reasoning length with only a 1.9\% increase in accuracy. These results suggest that while performance varies with model capacity, our method remains effective and adaptable across different model sizes. Furthermore, although the training data includes only math and code examples, the reasoning mode-switching capability of SABER generalizes well to unseen task types. Table~\ref{tab:main_results_7B} presents the performance of SABER on a logical reasoning benchmark, further demonstrating its strong generalization capabilities.

\subsection{Ablation Experiments}

To answer RQ3, we conduct ablation studies to assess the impact of key components in SABER, with each variant modifying exactly one component from the SABER configuration: (1) All Budget Downgrade: all training samples are downgraded by one reasoning level; (2) Without Budget Downgrade: all samples are trained strictly under their base model’s original reasoning budget; (3) Reduce NoThink Ratio: decrease the proportion of NoThink samples in training to 30\%; (4) Remove NoThink Data: remove all NoThink-mode data from training; and (5) Without Accuracy Filtering: during preprocessing, do not use model answer correctness to filter samples when assigning difficulty tiers.

As shown in Table~\ref{tab:ablation_result}, removing budget downgrade slows the adaptation to different reasoning modes, while applying aggressive downgrade to all samples leads to unstable training and reduced accuracy. Here, the “Without Budget Downgrade” setting closely resembles the baseline~\cite{li2025selfbudgeteradaptivetokenallocation}. Without progressive budget guidance, the model struggles to generalize shorter reasoning modes, resulting in slow mode adaptation and reduced efficiency. Reducing or removing NoThink samples significantly hurts NoThink performance without improving other modes, confirming the necessity of joint training. Finally, removing accuracy-based filtering causes supervision noise and degrades stability. These results underscore that all individual components of SABER are essential for effective and stable learning.

\subsection{Behaviors of Different Reasoning Modes}

To answer RQ4, we present a representative example from the MATH500 dataset to illustrate the differences among three thinking modes. As shown in Figure~\ref{fig_case_example}, all modes follow the core problem-solving steps: decomposing 196 and counting its divisors. These steps lead to the correct final answer, which is highlighted in green in the figure. Specifically, FastThink includes only these essential steps, making the reasoning process concise and direct. In contrast, CoreThink introduces some initial reasoning, such as “I remember that…”, before solving the problem. It also includes additional intermediate steps and occasional reflection during the decomposition process. Furthermore, DeepThink adopts a more thorough and reflective reasoning process. After reaching a solution, it includes justifications such as “I think that’s correct because…”, followed by a final double-check and self-verification. This demonstrates a deeper and more comprehensive reasoning pattern.

To further study what qualitative differences emerge among the switchable reasoning modes of SABER, we investigate the distribution of reflective markers, calculating the token‑level audit of cue words associated with verification (“check”, “verify”), retrospection (“recall”), branch exploration (“alternatively”), logical turns (“however”, “since”), and stepwise decomposition (“step‑by‑step”). As Figure~\ref{fig_freq} illustartes, FastThink generates significantly less reflective terms compared to DeepThink, with decreasement of 80.24\%, 70.83\%, and 61.97\% on GSM8K, MATH, MATH500 respectively. FastThink achieves the sharpest drop in reasoning depth, yet its answer accuracy remains almost unchanged. Thus, the policy of FastThink trims filler phrases while retaining transition tokens that signal genuine reasoning steps.

Additionally, we study the behaviors of different reasoning modes. Figure~\ref{fig_cot_len} illustrates the difference of three reasoning modes. Compareed with the DeepThink, CoreThink and FastThink cut the length of the chain‑of‑thought. Evaluated on GSM8K~\cite{Cobbe2021TrainingVT} based on SABER (Deepseek-R1-Distill-Qwen-7B), the average number of reasoning tokens of FastThink falls by 79\% relative to the DeepThink.

%% file: Sections/Conclusion.tex
\section{Conclusions and Future Work}
\label{Conclusion}

In this work, we present SABER, a unified and switchable reasoning framework that enables large language models to perform efficient and controllable reasoning across specific modes. By combining structured reward design, discrete reasoning modes, and curriculum-style budget assignment, SABER achieves high training stability and reasoning flexibility without requiring supervised warm start. Our experiments demonstrate that SABER generalizes well across math reasoning, code generation, and logical reasoning tasks, maintaining strong performance under varying computational constraints. We further show that SABER effectively supports both thinking and no-thinking modes within a single model, with minimal performance degradation. These results point towards a promising direction for controllable and cost-efficient reasoning in LLMs.

Despite the encouraging performance, several caveats remain. Scaling the framework to heterogeneous, real-world workloads may expose unforeseen practical hurdles that our controlled experiments do not cover. The pipeline also depends on an assessor LLM to calculate a reasoning budget for each prompt and on a subsequent manual grouping into difficulty bands. External variables such as alternative decoding strategies could also affect the transferability of the method. These considerations do not undermine the contribution, but rather mark clear directions for future investigation.